\documentclass[10pt,twocolumn,letterpaper]{article}

\usepackage[pagenumbers]{cvpr} 

\usepackage{amsmath}
\usepackage{graphicx}
\usepackage{booktabs}
\usepackage{multirow}
\usepackage{bm}
\usepackage{bbm}
\usepackage{soul}
\usepackage{makecell}
\usepackage{pifont}
\usepackage[breakable]{tcolorbox}
\usepackage[accsupp]{axessibility}
\usepackage{placeins}
\newcommand{\sysname}{SPARCL}
\newcommand{\longname}{\underline{S}ynthetic \underline{P}erturbations for \underline{A}dvancing \underline{R}obust \underline{C}ompositional \underline{L}earning}
\definecolor{cvprblue}{rgb}{0.21,0.49,0.74}
\usepackage[pagebackref,breaklinks,colorlinks,allcolors=cvprblue]{hyperref}

\def\paperID{11272} 
\def\confName{CVPR}
\def\confYear{2025}

\title{Enhancing Vision-Language Compositional Understanding with Multimodal Synthetic Data}

\author{Haoxin Li and Boyang Li
\\
{Nanyang Technological University}\\
\texttt{\small{\{haoxin003, boyang.li\}@ntu.edu.sg}}
}

\begin{document}
\maketitle
\begin{abstract}
Paired image-text data with subtle variations in-between (\eg, people holding surfboards vs. people holding shovels) hold the promise of producing Vision-Language Models with proper compositional understanding. Synthesizing such training data from generative models is a highly coveted prize due to the reduced cost of data collection. However, synthesizing training images for compositional learning presents three challenges: (1) efficiency in generating large quantities of images, (2) text alignment between the generated image and the caption in the exact place of the subtle change, and (3) image fidelity in ensuring sufficient similarity with the original real images in all other places. We propose \sysname{} (\underline{S}ynthetic \underline{P}erturbations for \underline{A}dvancing \underline{R}obust \underline{C}ompositional \underline{L}earning), which integrates image feature injection into a fast text-to-image generative model, followed by an image style transfer step, to meet the three challenges. Further, to cope with any residual issues of text alignment, we propose an adaptive margin loss to filter out potentially incorrect synthetic samples and focus the learning on informative hard samples. Evaluation on four compositional understanding benchmarks demonstrates that \sysname{} significantly improves the compositionality of CLIP, boosting the average accuracy of the CLIP base model by over 8\% across all benchmarks and outperforming state-of-the-art methods by 2\% on three benchmarks.
\end{abstract}    
\section{Introduction}
\label{sec:intro}
\begin{figure}[t]
\centering
\begin{subfigure}{\linewidth}
    \centering
    \includegraphics[width=\linewidth]{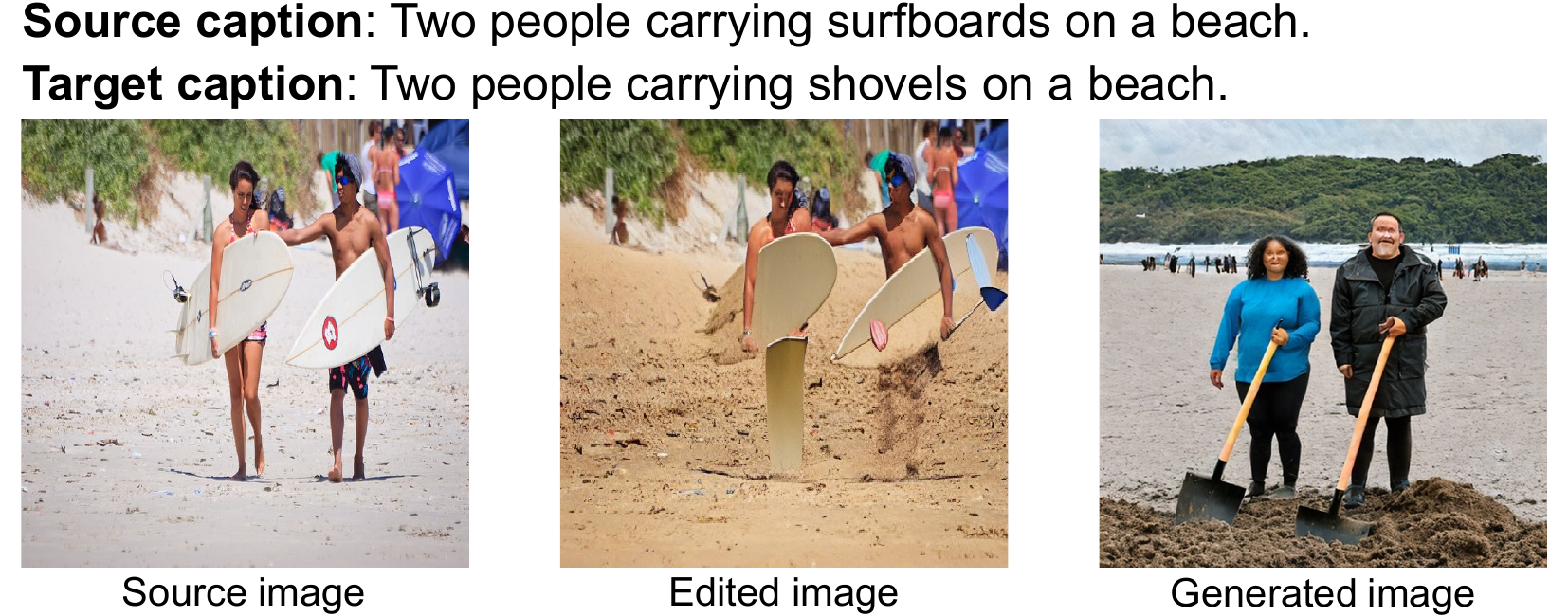}
    \caption{Difficulty in accurately creating precise variations.}
    \label{fig:challenges-1}
\end{subfigure}

\vspace{0.5em} 

\begin{subfigure}{\linewidth}
    \centering
    \includegraphics[width=\linewidth]{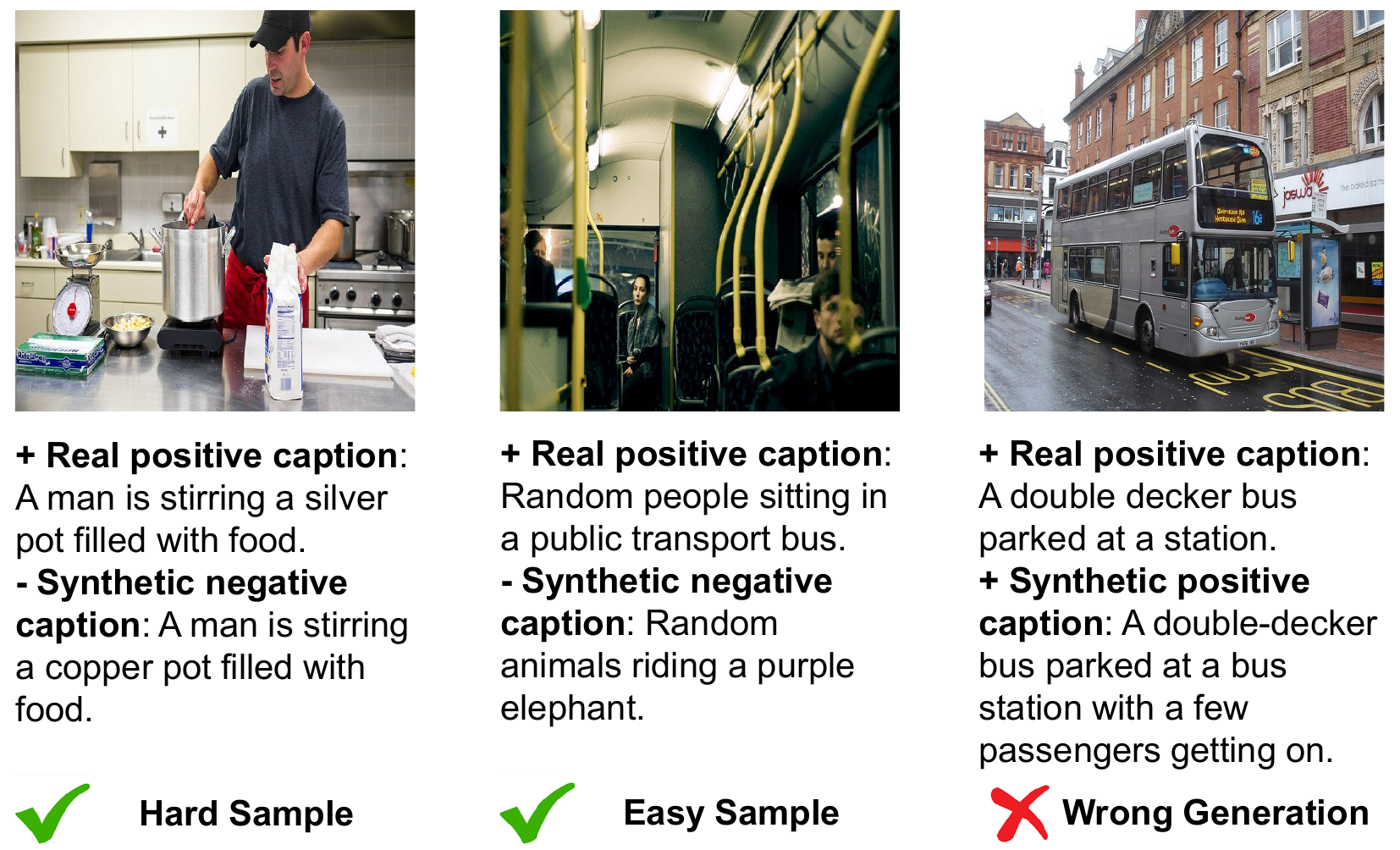}
    \caption{Inconsistency in cross-modal alignment quality of synthetic samples.}
    \label{fig:challenges-2}
\end{subfigure}

\caption{Challenges in generating and training on synthetic data: (a) When generating an image with subtle variations based on a real image and a target caption specifying the variations, an image editing model \cite{brooks2023instructpix2pix} struggles with text alignment (middle), while an image generation model \cite{rombach2022high} fails to maintain image fidelity (right). (b) Synthetic positive and negative image-caption pairs show different levels of alignment quality. The subtle variations in the synthetic negative caption (left) make it difficult to distinguish from the positive; the over-modified negative caption (middle) is easy to distinguish; and the hallucinated content in the synthetic positive caption (right) results in an incorrect positive.}
\label{fig:challenges}
\vspace{-1em}
\end{figure}

Current Vision-Language Models (VLMs) still face limitation in accurately interpreting compositional relationships between objects and attributes, as demonstrated by numerous evaluations \cite{thrush2022winoground,zhao2022vl,yuksekgonul2022and,ma2023crepe,hsieh2023sugarcrepe}. This limitation primarily stems from the absence of subtle variations in the training data \cite{kamath2023s} (\eg, the subtle  variations between the two captions in Figure \ref{fig:challenges} (a)). As a result, it becomes possible to maximize empirical image-caption alignment using shortcut features \cite{geirhos2020shortcut} rather than genuinely learning nuanced distinctions. While collecting training samples with subtle variations could enhance compositionality, this approach is time-consuming and labor-intensive, rendering it impractical at scale.

Advances in generative models \cite{brown2020language,zhang2022opt,touvron2023llama,karras2019style,chang2023muse,rombach2022high,saharia2022photorealistic,yu2022scaling,luo2023latent,esser2024scaling} now facilitate the synthetic generation of training samples with subtle variations between paired samples \cite{yuksekgonul2022and,doveh2023teaching,castro2024clove,singh2024learn,shou2024enhancing,oh2024preserving,zhang2023contrasting,sahin2024enhancing,peng2024synthesize,zhang2024countercurate,lai2024improving}. By starting with real image-caption pairs, generative models can create subtle edits to both captions and images, providing valuable training data with minimal manual effort. These generated variations enable VLMs to enhance their compositionality by learning from the nuanced differences.

However, generating and training on synthetic data presents two key challenges. The first is the difficulty of efficiently and accurately creating precise variations in synthetic images. To generate large-scale training data with precise variations, the image generation process must meet three criteria: \emph{efficiency} in producing large quantities of images, \emph{text alignment} of the generated image to the corresponding caption, which is subtly changed from the original caption, and \emph{image fidelity} in ensuring the generated image otherwise closely matches their real counterpart. Nevertheless, current image generation methods struggle to meet all three criteria simultaneously. Per-sample optimization methods \cite{kawar2023imagic,valevski2023unitune,zhang2023sine,shi2024dragdiffusion} lack efficiency, while zero-shot image editing methods \cite{meng2021sdedit,brooks2023instructpix2pix,garibi2024renoise} and text-to-image (T2I) models \cite{rombach2022high,saharia2022photorealistic} often fail to achieve proper text alignment and image fidelity, respectively, as illustrated in Figure \ref{fig:challenges} (a). The inaccurate synthetic variations produced by these models can mislead the learning of VLMs. To alleviate this issue, we take a fast T2I model, which excels in efficiency and text alignment, and inject real image features into the text prompt features in order to enhance image fidelity. By combining this approach with AdaIN \cite{huang2017arbitrary}, we manage to substantially improve the image generation process per the three criteria.

Still, the generated data do not \emph{perfectly} meet the text alignment criterion, and we propose to deal with the remaining problems with an innovative approach to model training. 
%
As Figure \ref{fig:challenges} (b) shows, the similarity between synthetic positive and negative pairs may vary, resulting in a mixture of hard text-image pairs, easy pairs, and incorrect pairs. Instead of treating all positive and negative samples uniformly, we propose a novel loss that differentiates between positive, hard negative, and easy negative samples. Further, we propose an adaptive margin that helps to filter out potentially incorrect generations and focus the learning on informative hard samples.

In summary, we propose \sysname{}, which integrates image feature injection into a fast T2I model to improve the quality of synthetic variations in images and employs an adaptive margin loss to leverage the varying alignment quality in synthetic samples for training VLMs. Evaluations on four compositional understanding benchmarks demonstrate that \sysname{} significantly enhances VLM compositionality, improving CLIP by over 5\% on VL-CheckList \cite{zhao2022vl} and 7\% on SugarCrepe \cite{hsieh2023sugarcrepe}, while surpassing state-of-the-art methods by 1\% and 2\% on the two benchmarks. The main contributions of this paper are as follows: (1) we propose image feature injection to enhance the quality of synthetic variations in images, which provide valuable training data to improve the compositionality of VLMs; (2) we introduce an adaptive margin loss to leverage varying levels of cross-modal alignment in synthetic samples to effectively differentiate positive and negative samples; (3) experimental results validate that \sysname{} significantly improves the compositional understanding capabilities of CLIP models.

\section{Related Work}
\subsection{Limitations in Compositionality of VLMs}
While VLMs excel in many multi-modal tasks \cite{radford2021learning,jia2021scaling,yao2021filip,singh2022flava,li2022blip,li2023blip,zhu2023minigpt}, they still struggle with compositional understanding---the ability to interpret novel combinations of known visual and textual components. Benchmarks like What’sUp \cite{kamath2023s} reveal difficulties in understanding spatial relationships, SPEC \cite{peng2023synthesize} highlights issues with object size, position, and count, and ARO \cite{yuksekgonul2022and} uncovers limitations in understanding attributes, relations, and word order. Winoground \cite{thrush2022winoground}, SNARE \cite{wang2023can}, and VL-CheckList \cite{zhao2022vl} also expose these shortcomings. SugarCrepe \cite{hsieh2023sugarcrepe} addresses hackable biases in prior benchmarks, where text-only models achieve artificially high performance, by introducing fluent and meaningful hard negatives. Their findings suggest that previous benchmarks overestimated compositional understanding. Building on this, SugarCrepe++ \cite{dumpala2024sugarcrepe++} further introduces semantically equivalent but lexically varied captions as hard positives and shows the difficulties of VLMs in distinguishing between lexical variations.

\subsection{Improving Compositionality of VLMs}
Prior approaches to improving the compositionality of VLMs can be broadly classified into the following categories:
\emph{(1) Leveraging detailed image captions:} Detailed captions from dense captioning models \cite{doveh2023dense,li2024desco}, simulation platforms \cite{cascante2023going}, and video annotations \cite{khan2024figclip} are collected to train VLMs. However, these samples often lack pairs with subtle variations, limiting their contribution to compositionality.
\emph{(2) Distilling from pretrained models:} SDS-CLIP \cite{basu2023augmenting}, SF-CLIP \cite{sameni2024building} and IL-CLIP \cite{zheng2024iterated} distill knowledge from pretrained image generation models and visual-language foundation models. IL-CLIP \cite{zheng2024iterated} refines representations through iterative learning with pretrained vision and language agents. However, pretrained models also face limitations in compositionality \cite{hua2024mmcomposition,tong2024eyes}.
\emph{(3) Incorporating structural knowledge:} MosaiCLIP \cite{singh2023coarse} and Structure-CLIP \cite{huang2024structure} incorporate scene graph knowledge in text features. CLIP-SGVL \cite{herzig2023incorporating} and 3VL \cite{yellinek20233vl} train VLMs to predict scene graphs. \cite{mitra2023compositional} utilizes scene graphs as prompts to elicit compositional knowledge from VLMs without further training. However, these methods rely on models trained with expensive dense structure annotations (\eg, scene graphs).
\emph{(4) Utilizing synthetic negative samples:} Rule-based tools \cite{honnibal2017spacy} or large language models (LLMs) \cite{devlin2018bert,radford2018improving} are used to generate negative captions by editing real captions \cite{yuksekgonul2022and,doveh2023teaching,castro2024clove,singh2024learn,shou2024enhancing,oh2024preserving,zhang2023contrasting,cascante2024natural}. Image generation models \cite{rombach2022high} are also used to create or edit images for training \cite{sahin2024enhancing,peng2024synthesize,zhang2024countercurate,lai2024improving}. Despite their utility, efficiently generating large amount of images with precise variations remains challenging due to the inherent limitations of generative models.
\emph{(5) Applying fine-grained alignment constraints:} MCD \cite{kim2023misalign} enforces multi-scale alignment across images with varying augmentations and the corresponding text captions. SPARC \cite{bica2024improving} learns local alignment by associating each text token with a group of local image patches. CE-CLIP \cite{zhang2023contrasting} applies intra-modal contrastive loss and cross-modal ranking loss to improve alignment. However, uniform supervision applied to synthetic samples with varying alignment quality limits their effectiveness.
In this paper, we propose \sysname{} to address two key challenges in learning from synthetic data: the difficulty in accurately creating precise variations and the inconsistency in cross-modal alignment quality in synthetic data.

\subsection{Training with Synthetic Data}
Synthetic data for training machine learning models have been studied in various fields \cite{nikolenko2021synthetic,rosenberg2019speech,rossenbach2020generating,kumar2020data,yang2020generative,he2022generate,meng2022generating,dan2020generative,tucker2020generating}. Synthetic data generated through simulations and graphics engines supports a wide range of tasks \cite{dosovitskiy2015flownet,peng2017visda,varol2021synthetic}. However, these synthetic datasets often diverge significantly from real-world data. Recent advances in generative models \cite{brown2020language,zhang2022opt,touvron2023llama,karras2019style,chang2023muse,rombach2022high,saharia2022photorealistic,yu2022scaling,luo2023latent} have made it possible to synthesize data that more closely resembles real-world scenarios. Synthetic data are widely used in both language tasks \cite{gao2023self,honovich2022unnatural,meng2022generating,meng2023tuning,wang2022self,west2021symbolic,zhu2022visualize,tang2023learning,yang2022z,lu2022imagination} and vision tasks \cite{besnier2020dataset,he2022synthetic,azizi2023synthetic,sariyildiz2023fake,hinterstoisser2018pre,tremblay2018training,zhang2021datasetgan,jahanian2021generative,liu2022palm,baradad2021learning,tian2024stablerep}. Synthetic data are often noisy, necessitating noise-resistant training methods \cite{jiang2018mentornet,li2019learning,chen2019understanding,yao2020searching,li2021learning,huang2023twin}, especially those designed for contrastive learning \cite{liu2021noise,Ibrahimi_2022_WACV}. In this paper, we generate multimodal samples with subtle variations and filter out potentially incorrect ones during training to improve the compositionality of VLMs.

\begin{figure*}[ht]
\centering
\includegraphics[width=\linewidth]{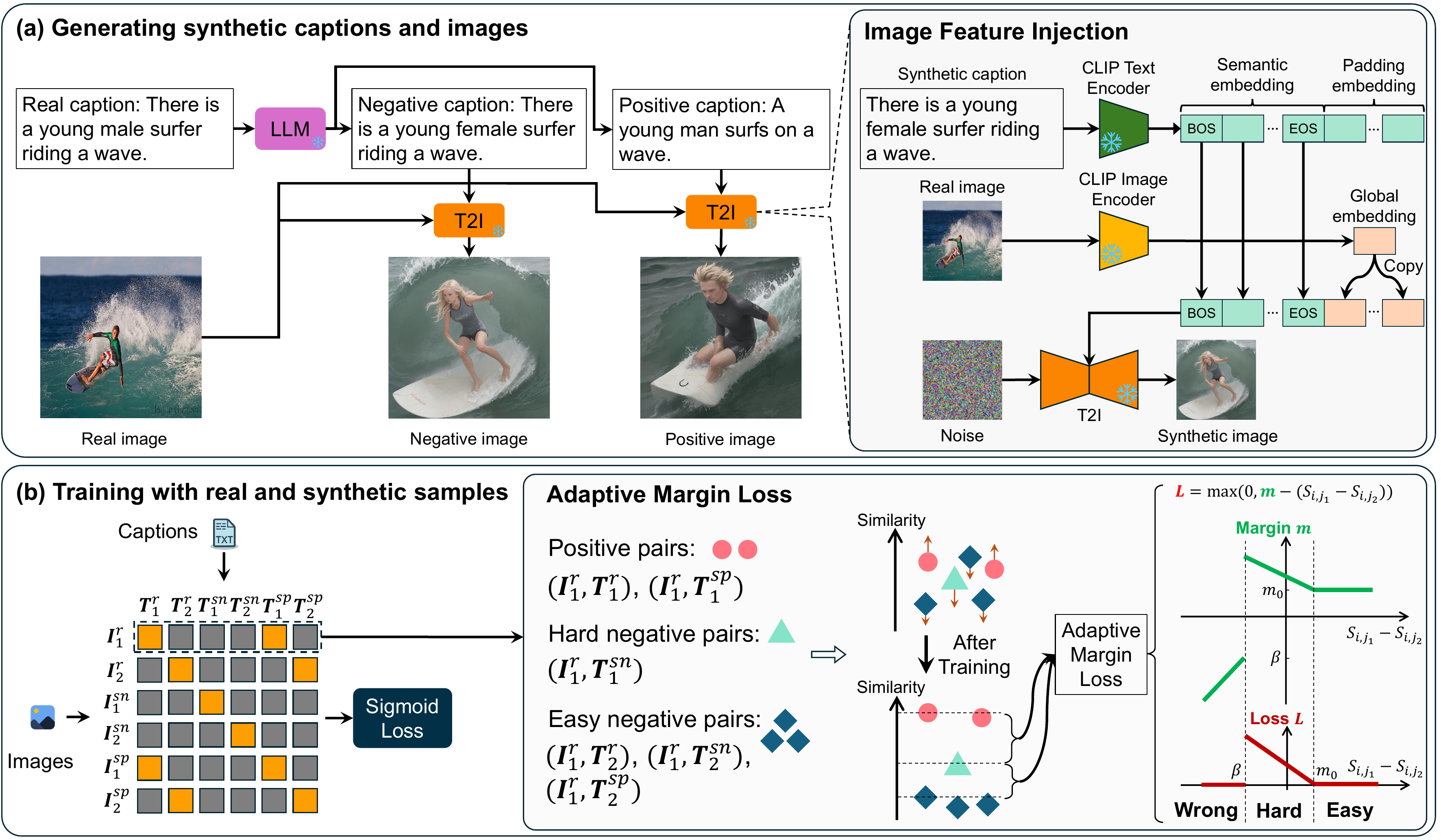}
\caption{An overview of \sysname{}. (a) Starting with a real image-caption pair, we generate synthetic positive and negative pairs with subtle variations using an LLM and a fast T2I model. To improve the quality of subtle variations in synthetic images, we introduce image feature injection to reduce unintended variations from a standard T2I model (see Sec. \ref{subsec:feat-injection}). (b) We train the VLM using both real and synthetic samples. In addition to a sigmoid loss for distinguishing positive and negative pairs, we apply an adaptive margin loss that leverages varying alignment levels across training samples to learn informative nuanced distinctions (see Sec. \ref{subsec:loss}).}
\label{fig:framework}
\vspace{-1em}
\end{figure*}

\section{\sysname{}}
To improve the compositional understanding abilities of VLMs, we propose \longname{}, or \sysname{}, which generates multimodal samples with subtle variations from real samples and trains VLMs to learn nuanced differences through synthetic data. In the generation phase, \sysname{} creates positive and negative captions with slight variations from real captions using an LLM, then generates images based on the real image and these modified captions through a fast T2I model. To enhance the quality of subtle variations in synthetic images, we introduce image feature injection, which integrates real image features into the text prompt features of the T2I model to improve fidelity to the real image, as detailed in Sec. \ref{subsec:feat-injection}. In the training phase, \sysname{} employs an adaptive margin loss that leverages varying levels of multimodal alignment in the synthetic samples to effectively learn informative nuanced distinctions, as described in Sec. \ref{subsec:loss}. The framework of \sysname{} is shown in Figure \ref{fig:framework}.

\subsection{Generating Negative and Positive Captions}
Captions with subtle variations are crucial for learning compositional knowledge, as shown by previous work that generates captions by randomly swapping or replacing nouns and adjectives \cite{yuksekgonul2022and,zhang2023contrasting}. However, manually designed generation rules often introduce nonsensical or grammatically incorrect artifacts, creating shortcut features \cite{geirhos2020shortcut} that obstruct true compositional understanding \cite{hsieh2023sugarcrepe}. To address this issue, we use an LLM to generate natural synthetic captions. To further mitigate the impact of generative artifacts, we generate both negative and positive captions, ensuring that VLMs cannot easily differentiate negative captions from positive ones based solely on artifacts. We denote the $i^{th}$ real image-caption pair in the training set as $(\bm{I}_i^{r}, \bm{T}_i^{r})$. Given a real caption $\bm{T}_i^{r}$, we prompt the LLM to generate a synthetic negative caption $\bm{T}_i^{sn}$ and a synthetic positive caption $\bm{T}_i^{sp}$, as specified by the prompts in Figure \ref{fig:prompts} in the Appendix.

\subsection{Generating Images via Image Feature Injection}
\label{subsec:feat-injection}
Synthetic images that exhibit subtle variations from real images while aligning with the captions are valuable but challenging to generate, as discussed in Sec. \ref{sec:intro}. Although previous works \cite{sahin2024enhancing,peng2024synthesize,zhang2024countercurate,lai2024improving} have utilized object segmentation or filtered dissimilar generations to improve fidelity to real images, they struggle with manipulating relationships or lack efficiency. We aim to enhance fidelity by injecting image features into a fast T2I model \cite{luo2023latent}, which already achieves high efficiency and text alignment.

\vspace{0.5em}\noindent\textbf{Image Feature Injection.} The images generated by T2I models lacks image fidelity to real images, as no real image information is input to the models. To enhance the fidelity of synthetic images, we inject real image features into the text prompt features, enabling the model to incorporate information from the real images. 

In T2I models, content and style are separated in the semantic and padding embeddings. The semantic embeddings (before the embedding of the [\text{EOS}] token) usually capture most of the image content in the text prompts, while the padding embeddings (after the [\text{EOS}] token) usually represent the image style \cite{yu2024uncovering}. Therefore, we can inject real image features into the padding embeddings to guide the model in generating images with a similar style to real images, without affecting the alignment with the captions, thanks to the decoupling of content and style in the prompt embeddings.

We extract features for a real image $\bm{I}_i^{r}$ and the corresponding synthetic caption $\bm{T}_i^{s}$ ($s\in\{sn, sp\}$) using two aligned feature encoders (\eg, CLIP image and text encoders). The image embedding $\bm{f}_i^{r}$ is the encoder output at the position of the [\text{CLS}] token. The text embedding is the text encoder output $\bm{e}_i^{s}=\langle\bm{e}_{i,j}^{s}\rangle_{j=1}^{L}$, where $L$ is the maximum sequence length. Let the index of the [\text{EOS}] token in $\bm{e}_i^{s}$ be $k_i^{s}$, $k_i^{s} \le L$. That is, the padding token [PAD] is used as input to the text encoder at positions between $k_i^{s}$ and $L$. The key step of \sysname{} is to replace the text embeddings at those positions with the image embedding, producing a new sequence of text embeddings, $\hat{\bm{e}}_i^{s}$,
\begin{small}
\begin{equation}
\label{eq:img_inj}
    \hat{\bm{e}}_i^{s} = \langle \bm{e}_{i,1}^{s}, \bm{e}_{i,2}^{s}, \ldots, \bm{e}_{i,k_i}^{s}, \underbrace{\bm{f}_i^{r}, \ldots, \bm{f}_i^{r}}_{L-k_i^{s} \text{ times}}\rangle.
\end{equation}
\end{small}%
The embeddings $\hat{\bm{e}}_i^{s}$ is then used as input to the T2I model to generate a synthetic image $\tilde{\bm{I}}_i^{s}$. This process, depicted in Figure \ref{fig:framework}, reduces unintended variations in synthetic images generated by the standard T2I model and enhances image fidelity to real images, thereby improving the quality of synthetic subtle variations.

\vspace{0.5em}\noindent\textbf{Style Transfer.} To further reduce the domain gap between synthetic and real images, we use AdaIN \cite{huang2017arbitrary} to transfer the style of the synthetic image to that of the real image. 
We use a pretrained AdaIN encoder to extract content features from $\tilde{\bm{I}}_i^{s}$ and style features from $\bm{I}_i^{r}$. We normalize the content features using instance normalization and then scale and shift them with the mean and variance of the style features. The transformed features are fed into a pretrained AdaIN decoder to generate $\bm{I}_i^{s}$, which is subsequently used for model training.

\subsection{Training with Real and Synthetic Samples}
\label{subsec:loss}
After generating captions and images, each real image-caption pair $(\bm{I}_i^{r}, \bm{T}_i^{r})$ is extended to $(\bm{I}_i^{r}, \bm{T}_i^{r}, \bm{I}_i^{sn}, \bm{T}_i^{sn}, \bm{I}_i^{sp}, \bm{T}_i^{sp})$ by adding one synthetic negative and one synthetic positive pair. We then train the VLM using these extended samples. Given a batch of $n$ sample groups, $\{(\bm{I}_i^{r}, \bm{T}_i^{r}, \bm{I}_i^{sn}, \bm{T}_i^{sn}, \bm{I}_i^{sp}, \bm{T}_i^{sp})\}_{i=1}^n$, we organize the images by concatenating all real, synthetic negative, and synthetic positive images into an image batch $\bm{I}^{B}$ of $3n$ images, and similar all captions into a caption batch  $\bm{T}^{B}$ of $3n$ captions.


We calculate the similarity between each image-caption pair in the batch. The similarity between the $i^{\text{th}}$ image $\bm{I}_i^{B}$ and the $j^{\text{th}}$ caption $\bm{T}_j^{B}$ is given by $S_{i,j} = c(\mathcal{E}_I(\bm{I}_i^{B}), \mathcal{E}_T(\bm{T}_j^{B}))$, where $c(\cdot, \cdot)$ denotes cosine similarity, and $\mathcal{E}_I$ and $\mathcal{E}_T$ represent the image and text encoders of the VLM, respectively. We then define the ground-truth alignment variable $M_{i,j}$, which takes value $1$ if the image matches with the caption, and $-1$ otherwise. 
Clearly, the only positive image-caption pairs are $(\bm{I}_i^{r}, \bm{T}_i^{r}), (\bm{I}_i^{sn}, \bm{T}_i^{sn}), (\bm{I}_i^{sp}, \bm{T}_i^{sp}), (\bm{I}_i^{r}, \bm{T}_i^{sp})$ and $(\bm{I}_i^{sp}, \bm{T}_i^{r})$. 
%
To account for multiple positive associations of an image or a caption, we apply a sigmoid-based contrastive loss \cite{zhai2023sigmoid,bulat2024fff} to encourage higher similarity for positive pairs and lower similarity for negative pairs,
\begin{small}
\begin{equation}\label{eq:contr-loss}
    L_{con}=-\frac{1}{3n}\sum_{i=1}^{3n}\sum_{j=1}^{3n}{\text{log}\frac{1}{1+\text{exp}(-M_{i,j}(S_{i,j}/\tau+b))}},
\end{equation}
\end{small}%
where $\tau$ is the temperature parameter and $b$ is a bias term.

\vspace{0.5em}\noindent\textbf{Adaptive Margin Loss.} Despite best efforts toward controlling generative models, synthetic samples may still have variable quality in terms of the alignment between text and imagery. That is, purported positive (resp. negative) pairs may not be semantically similar (resp. dissimilar). To account for variation in synthetic data quality, we propose to differentiate between real data and synthetic data and between easy negatives and hard negatives. Further, we propose 
an adaptive margin that filters out potential incorrect samples and prioritizes learning from hard samples.


For each image in a batch, we define four sets of captions, a positive set $\mathbb{P}$, a hard negative set $\mathbb{N}_h$, an easy negative set $\mathbb{N}_e$, and a real negative set $\mathbb{N}_r$, which represent different levels of alignment within the batch. If the image is a real or synthetic positive image,
\begin{itemize}
\item $\mathbb{P}$ contains the real positive captions $\bm{T}_i^{r}$ and the synthetic positive captions $\bm{T}_i^{sp}$ of the current, $i$-th image.
\item $\mathbb{N}_h$ contains the synthetic negative captions $\bm{T}_i^{sn}$ of the current image.
\item $\mathbb{N}_e = \{\bm{T}_j^{r}, \bm{T}_j^{sp}, \bm{T}_j^{sn}| i \neq j \}$ contains all real and synthetic captions from all other images.
\item $\mathbb{N}_r = \{\bm{T}_j^{r}| i \neq j \}$ contains all real captions belonging to the other images. 
\end{itemize}
If the image is a synthetic negative image, the sets $\mathbb{N}_e$ and $\mathbb{N}_r$ are unchanged but $\mathbb{P}$ and $\mathbb{N}_h$ differ:
\begin{itemize}
\item $\mathbb{P}$ contains the synthetic negative captions $\bm{T}_i^{sn}$. 
\item $\mathbb{N}_h$ contains $\bm{T}_i^{r}$ and $\bm{T}_i^{sp}$. 
\end{itemize}
%
%
We define the margin loss for the $i^{th}$ image as:
\begin{footnotesize}
\begin{equation}\label{eq:ada-margin-loss}
\begin{split}
    L_{mar,i}^{I}&=\frac{1}{\lvert\mathbb{P}\rvert\cdot\lvert\mathbb{N}_h\rvert}\sum_{j_1\in\mathbb{P},j_2\in\mathbb{N}_h}\text{max}(0, m+S_{i,j_2}-S_{i,j_1}) \\
    &+\frac{1}{\lvert\mathbb{N}_h\rvert\cdot\lvert\mathbb{N}_e\rvert}\sum_{j_1\in\mathbb{N}_h,j_2\in\mathbb{N}_e}\text{max}(0, m+S_{i,j_2}-S_{i,j_1}) \\
    &+\frac{\alpha}{\lvert\mathbb{P}\rvert\cdot\lvert\mathbb{N}_r\rvert}\sum_{j_1\in\mathbb{P},j_2\in\mathbb{N}_r}\text{max}(0, m+S_{i,j_2}-S_{i,j_1}).
\end{split}
\end{equation}
\end{footnotesize}%
This loss encourages positive pairs to have higher similarity scores than hard negative pairs, which should have higher similarity than easy negative pairs. Additionally, a weight $\alpha > 1$ is applied to comparisons between positive pairs and real negative pairs to emphasize these comparisons, as real samples are generally correct, making comparisons involving them more reliable.

Lastly, we propose an adaptive margin in the above loss. Let $d=S_{i,j_1}-S_{i,j_2}$ denote the difference between two similarity scores in Eq. \eqref{eq:ada-margin-loss}. The adaptive margin $m$ is then computed as follows:
\begin{small}
\begin{equation}\label{eq:ada-margin}
    m=\left\{
        \begin{array}{ll}
            d, & d<\beta \\
            (\frac{m_0-d}{m_0-\beta}\gamma+1)m_0, & \beta\leq d\leq m_0 \\
            m_0 & d>m_0
        \end{array}
    \right.
\end{equation}
\end{small}%
where $m_0$ is a base margin, $\beta<0$ is a cutoff threshold, $\gamma$ is a scaling factor. The adaptive margin is designed with the following rationale: when $d<\beta$, the purported positive pair is much less similar than the purported negative pair, suggesting the presence of incorrect or mislabeled samples, so we zero out the loss by setting the margin to $d$. For $\beta\leq d\leq m_0$, the margin is scaled, with smaller differences receiving larger margins to emphasize learning from harder samples. When $d>m_0$, the samples are already well-separated with the margin $m_0$, so the margin is capped at $m_0$, again leading to zero loss. We visualize both the adaptive margin and the corresponding loss as functions of $d$ in Figure \ref{fig:framework} (b).

The loss for all images is given by $L_{mar}^{I}=\frac{1}{3n}\sum_{i}L_{mar,i}^{I}$. Analogously, we compute $L_{mar}^{T}$ for all captions. The total adaptive margin loss is $L_{mar}=L_{mar}^{I}+L_{mar}^{T}$. Finally, the overall training loss is a weighted combination of the contrastive loss and the adaptive margin loss, with weight $\lambda$:
\begin{small}
\begin{equation}
    L=L_{con}+\lambda L_{mar}.
\end{equation}
\end{small}%

\begin{table*}[t]
\begin{center}
\caption{Comparison of accuracy (\%) between \sysname{} and baselines on four benchmarks. ``img.'' represents images, ``cap.'' represents captions, ``syn.'' represents synthetic data.}
\label{tab:res-all}
\footnotesize
\begin{tabular}{@{}lp{0.1cm}cccccp{0.1cm}cccc@{}}
\toprule
\multirow{3}{*}{Method} && \multicolumn{5}{c}{Training Data} && \multirow{3}{*}{ARO} & \multirow{3}{*}{VL-CheckList} & \multirow{3}{*}{SugarCrepe} & \multirow{3}{*}{SugarCrepe++} \\
\cmidrule{3-7}
~ && Source & \makecell{\# real\\img.} & \makecell{\# real\\cap.} & \makecell{\# syn.\\img.} & \makecell{\# syn.\\cap.} && ~ & ~ & ~ & ~ \\
\midrule
CLIP \cite{radford2021learning} (Zero-Shot) && - & - & - & - & - && 61.1 & 73.2 & 73.4 & 59.8 \\
\midrule
CLIP \cite{radford2021learning} (Finetune) && COCO & 82K & 410K & 0 & 0 && 64.1 & 72.8 & 79.9 & 62.3 \\
SDS-CLIP \cite{basu2023augmenting} && COCO & 82K & 410K & 0 & 0 && 57.5 & - & - & - \\
\cite{sahin2024enhancing} && COCO & 0 & 0 & 82K & 82K && 65.0 & 69.9 & - & - \\
AMR-NegCLIP \cite{shou2024enhancing} && COCO & 100K & 100K & 0 & 500K && 79.4 & - & 85.2 & - \\
NegCLIP \cite{yuksekgonul2022and} && COCO & 100K & 100K & 0 & 500K && 76.0 & 74.6 & 82.5 & 64.9 \\
MosaiCLIP \cite{singh2023coarse} && COCO & 109K & 109K & 0 & 981K && \textbf{80.3} & 76.8 & - & - \\
FSC-CLIP \cite{oh2024preserving} && COCO & 100K & 100K & 0 & 1.5M && - & 77.2 & 85.1 & - \\
CE-CLIP \cite{zhang2023contrasting} && COCO & 82K & 410K & 0 & 2M && 79.7 & 76.3 & 85.2 & - \\
COMO \cite{lai2024improving} && COCO & 113K & 567K & 567K & 567K && - & 76.9 & - & - \\
\midrule
\sysname{} (our method) && COCO & 82K & 410K & 820K & 820K && 77.2 & \textbf{79.2} & \textbf{87.1} & \textbf{66.1} \\
\midrule
SPEC \cite{peng2023synthesize} && LAION & 20K & 20K & 20K & 20K && 70.1 & - & - & - \\
CounterCurate \cite{zhang2024countercurate} && Flickr & 30K & 30K & 150K & 150K && - & - & 82.8 & - \\
\cite{doveh2023teaching} && CC3M & 3M & 3M & 0 & 9M && - & 75.3 & - & 55.3 \\
CE-CLIP+ \cite{zhang2023contrasting} && COCO+CC3M & 3M & 3M & 0 & 15M && 80.4 & 79.3 & 87.5 & - \\
CLOVE \cite{castro2024clove} && LAION-COCO & $>$1B & $>$1B & 0 & $>$1B && 73.2 & - & 85.1 & - \\
IL-CLIP \cite{zheng2024iterated} && CC12M & 12M & 12M & 0 & 0 && - & - & 70.3 & - \\
SF-CLIP \cite{sameni2024building} && YFCC15M & 15M & 15M & 0 & 0 && - & - & 71.2 & - \\
syn-CLIP \cite{cascante2023going} && SyViC & 0 & 0 & $>$1M & $>$1M && 69.2 & 74.8 & - & - \\
FiGCLIP \cite{khan2024figclip} && VidSitu & \multicolumn{2}{c}{20K videos} & 0 & 0 && 67.0 & - & 74.6 & - \\
\bottomrule
\end{tabular}
\end{center}
\vspace{-1em}
\end{table*}

\section{Experiments}
\subsection{Datasets}
\noindent\textbf{Training.} We use the COCO-2014 dataset \cite{lin2014microsoft} as the training data source. The training set consists of 82,783 images, each paired with five captions. For each real image-caption pair, we generate one positive and one negative synthetic pair. In line with previous approaches \cite{doveh2023teaching, yuksekgonul2022and, zhang2023contrasting}, we train the VLMs using both the original COCO-2014 training data and the synthetic samples.

\vspace{0.5em}\noindent\textbf{Evaluation.} We evaluate our model on four vision-language compositional understanding benchmarks: \emph{(1) ARO} \cite{yuksekgonul2022and}, which consists of 23,937 cases for relation understanding and 28,748 for attribute understanding. We exclude the subsets for order understanding, as they contain significant nonsensical and non-fluent artifacts \cite{hsieh2023sugarcrepe}. \emph{(2) VL-CheckList} \cite{zhao2022vl}, a large-scale benchmark with over 100,000 samples, evaluates compositionality across subsets of objects, attributes, and relationships, which are further divided into various fine-grained categories. \emph{(3) SugarCrepe} \cite{hsieh2023sugarcrepe} includes 7,000 test cases across seven subsets. In the above three benchmarks, each test case containing one image, one positive caption, and one negative caption. \emph{(4) SugarCrepe++} \cite{dumpala2024sugarcrepe++} includes 4,757 test samples across five subsets, where each test case consists of one image, two positive captions, and one negative caption. All benchmarks involve classifying captions as positive or negative for the given images. We report the average accuracy across all subsets of each benchmark and include the accuracy for each subset in Appendix \ref{supp-sec:res}.

\subsection{Implementation Details}
\noindent\textbf{Models.} We use ViT-B/32 and ViT-L/14 architectures from OpenAI’s CLIP model \cite{radford2021learning} as our base models, initialized with pretrained checkpoints. Following syn-CLIP \cite{cascante2023going}, we integrate LoRA adapters \cite{hu2021lora} into both the image and text encoders of CLIP to improve training efficiency and mitigate knowledge forgetting. Only the LoRA adapters are fine-tuned during training.

\vspace{0.5em}\noindent\textbf{Training Setups.} We use the AdamW optimizer \cite{loshchilov2017decoupled} with a cosine learning rate schedule \cite{loshchilov2016sgdr}. Training is conducted on two Tesla V100 GPUs, with a batch size of 128 sample groups for ViT-B/32 and 16 for ViT-L/14. The base learning rate is set to 0.01 for a total batch size of 256, and scaled linearly \cite{goyal2017accurate} based on the actual batch size. Training is performed for 3,000 steps for ViT-B/32 and 15,000 steps for ViT-L/14, corresponding to fewer than 5 epochs in previous studies \cite{doveh2023teaching,yuksekgonul2022and,zhang2023contrasting}. More details and hyperparameters settings are in Appendix \ref{supp-sec:exp}.

\begin{table*}[t]
\setlength{\tabcolsep}{3pt}
\begin{center}
\caption{Ablated performance (\%) of \sysname{}. ``SynCap'' refers to synthetic captions, ``SynImg'' refers to synthetic images, ``FeatInj'' denotes image feature injection, and ``CompSet'' indicates the comparison sets used in Eq. \eqref{eq:ada-margin-loss}.}
\label{tab:abl-res-all}
\scriptsize
\begin{tabular}{@{}cp{0.001cm}cp{0.001cm}cp{0.001cm}cccp{0.001cm}ccp{0.001cm}cccccp{0.001cm}@{}}
\toprule
\multirow{2}{*}{Model} && \multirow{2}{*}{Variant} && \multirow{2}{*}{SynCap} && \multirow{2}{*}{SynImg} & \multirow{2}{*}{FeatInj} & \multirow{2}{*}{AdaIN} && \multicolumn{2}{c}{Adaptive Margin Loss} && \multirow{2}{*}{ARO} & \multirow{2}{*}{VL-CheckList} & \multirow{2}{*}{SugarCrepe} & \multirow{2}{*}{SugarCrepe++} & \multirow{2}{*}{Average} \\
\cmidrule{11-12}
~ && ~ && ~ && ~ & ~ & ~ && CompSet & Margin && ~ & ~ & ~ & ~ & ~ \\
\midrule
\multirow{13}{*}{ViT-B/32} && \#1 && \ding{56} && \ding{56} & \ding{56} & \ding{56} && \multicolumn{2}{c}{\ding{56}} && 60.49 & 72.61 & 79.36 & 64.85 & 69.33 \\
\cmidrule{3-18}
~ && \#2 && \ding{52} && \ding{56} & \ding{56} & \ding{56} && \multicolumn{2}{c}{\ding{56}} && 71.77 & 73.52 & 86.35 & 64.32 & 73.99 \\
~ && \#3 && \ding{56} && \ding{52} & \ding{56} & \ding{56} && \multicolumn{2}{c}{\ding{56}} && 62.62 & 71.70 & 79.97 & 64.84 & 69.79 \\
~ && \#4 && \ding{52} && \ding{52} & \ding{56} & \ding{56} && \multicolumn{2}{c}{\ding{56}} && 71.86 & 75.54 & 85.43 & 65.22 & 74.51 \\
\cmidrule{3-18}
~ && \#5 && \ding{52} && \ding{52} & \ding{52} & \ding{56} && \multicolumn{2}{c}{\ding{56}} && 73.40 & 76.56 & 85.54 & 65.78 & 75.32 \\
~ && \#6 && \ding{52} && \ding{52} & \ding{56} & \ding{52} && \multicolumn{2}{c}{\ding{56}} && 73.79 & 75.72 & 85.79 & 64.49 & 74.95 \\
~ && \#7 && \ding{52} && \ding{52} & \ding{52} & \ding{52} && \multicolumn{2}{c}{\ding{56}} && 74.12 & 76.35 & 85.40 & 66.44 & 75.58 \\
\cmidrule{3-18}
~ && \#8 && \ding{52} && \ding{52} & \ding{52} & \ding{52} && All & Fixed && 76.79 & 78.59 & 87.08 & 65.42 & 76.97 \\
~ && \#9 && \ding{52} && \ding{52} & \ding{52} & \ding{52} && All & Adaptive && 77.15 & 79.16 & 87.11 & 66.12 & 77.38 \\
~ && \#10 && \ding{52} && \ding{52} & \ding{52} & \ding{52} && All & Adaptive Inversed && 76.67 & 79.26 & 86.68 & 65.30 & 76.97 \\
~ && \#11 && \ding{52} && \ding{52} & \ding{52} & \ding{52} && Only $(\mathbb{P},\mathbb{N}_h)$ & Adaptive && 77.48 & 80.48 & 86.15 & 64.70 & 77.20 \\
\midrule
\multirow{13}{*}{ViT-L/14} && \#1 && \ding{56} && \ding{56} & \ding{56} & \ding{56} && \multicolumn{2}{c}{\ding{56}} && 59.80 & 73.26 & 81.49 & 64.90 & 69.86 \\
\cmidrule{3-18}
~ && \#2 && \ding{52} && \ding{56} & \ding{56} & \ding{56} && \multicolumn{2}{c}{\ding{56}} && 72.16 & 75.44 & 87.38 & 64.44 & 74.85 \\
~ && \#3 && \ding{56} && \ding{52} & \ding{56} & \ding{56} && \multicolumn{2}{c}{\ding{56}} && 60.70 & 72.23 & 82.58 & 66.63 & 70.54 \\
~ && \#4 && \ding{52} && \ding{52} & \ding{56} & \ding{56} && \multicolumn{2}{c}{\ding{56}} && 75.20 & 78.29 & 87.29 & 64.61 & 76.35 \\
\cmidrule{3-18}
~ && \#5 && \ding{52} && \ding{52} & \ding{52} & \ding{56} && \multicolumn{2}{c}{\ding{56}} && 75.11 & 78.58 & 87.54 & 64.57 & 76.45 \\
~ && \#6 && \ding{52} && \ding{52} & \ding{56} & \ding{52} && \multicolumn{2}{c}{\ding{56}} && 74.88 & 79.10 & 87.50 & 64.62 & 76.52 \\
~ && \#7 && \ding{52} && \ding{52} & \ding{52} & \ding{52} && \multicolumn{2}{c}{\ding{56}} && 74.93 & 80.04 & 87.79 & 65.30 & 77.01 \\
\cmidrule{3-18}
~ && \#8 && \ding{52} && \ding{52} & \ding{52} & \ding{52} && All & Fixed && 75.21 & 80.75 & 88.02 & 66.41 & 77.60 \\
~ && \#9 && \ding{52} && \ding{52} & \ding{52} & \ding{52} && All & Adaptive && 75.83 & 80.81 & 88.23 & 66.83 & 77.93 \\
~ && \#10 && \ding{52} && \ding{52} & \ding{52} & \ding{52} && All & Adaptive Inversed && 75.15 & 80.71 & 87.93 & 66.61 & 77.60 \\
~ && \#11 && \ding{52} && \ding{52} & \ding{52} & \ding{52} && Only $(\mathbb{P},\mathbb{N}_h)$ & Adaptive && 76.26 & 80.57 & 87.42 & 66.28 & 77.63 \\
\bottomrule
\end{tabular}
\end{center}
\vspace{-1em}
\end{table*}

\subsection{Main Results}
We compare \sysname{} with several baseline methods: (1) methods that utilize synthetic samples for training \cite{lai2024improving,oh2024preserving,sahin2024enhancing,yuksekgonul2022and,shou2024enhancing,zhang2023contrasting,zhang2024countercurate,peng2023synthesize}, (2) methods that distill knowledge from pretrained models \cite{basu2023augmenting,zheng2024iterated,sameni2024building,castro2024clove}, (3) methods that incorporate knowledge from scene graphs \cite{singh2023coarse,doveh2023teaching}, and (4) methods that leverage detailed image captions \cite{cascante2023going,khan2024figclip}. Both \sysname{} and the baseline methods use ViT-B/32 as the base model. The results are presented in Table \ref{tab:res-all}.

We observe that \sysname{} achieves the best performance on VL-CheckList, SugarCrepe, and SugarCrepe++ compared to baselines trained on the same data source, namely COCO. Specifically, \sysname{} surpasses the strongest baseline by 1.4\% on VL-CheckList and 2.5\% on SugarCrepe, underscoring its effectiveness in enhancing compositional understanding. However, \sysname{} performs worse on ARO than some baselines. We hypothesize that this may be due to nonsensical or grammatically incorrect artifacts in the ARO test samples \cite{hsieh2023sugarcrepe}, which could favor methods that utilize rule-based synthetic training samples (\eg, \cite{singh2023coarse}) containing similar artifacts. Notably, even with only 50\% of the training data, \sysname{} outperforms most baseline methods (see Table \ref{tab:abl-res-sample} for \sysname{} performance with reduced training data). Compared to baselines using additional data sources, \sysname{} surpasses or matches their performance. For example, \sysname{} outperforms methods in \cite{doveh2023teaching,castro2024clove} despite their use of more training data. \sysname{} performs comparably to CE-CLIP+ \cite{zhang2023contrasting} on VL-CheckList and SugarCrepe, despite CE-CLIP+ leveraging CC3M as an additional data source and utilizing significantly more samples.

\subsection{Ablation Study and Analysis}
\label{subsec:ablation}
We perform ablation studies to assess the impact of each component and design choice in \sysname{}. The results, presented in Table \ref{tab:abl-res-all}, show the overall performance. 

\vspace{0.5em}\noindent\textbf{Synthetic captions vs. synthetic images.} To analyze the impact of synthetic captions and images, we compare Variant \#1 (using only real samples) with Variant \#2 (using real samples and synthetic captions) and \#3 (using real samples and synthetic images). Variant \#2 shows a significant improvement over Variant \#1, with the average accuracy improving from 69.33\% to 73.99\% for ViT-B/32. However, Variant \#3 obtains only a marginal improvements of about 0.5\%. These results suggest that synthetic captions substantially enhance compositional understanding, while synthetic images alone provide limited benefit, which aligns with findings in \cite{patel2024tripletclip}. We hypothesize that generative artifacts in synthetic images are more pronounced than in captions, which negatively affects the effective learning of nuanced distinctions by VLMs. When both synthetic captions and images are used (Variant \#4), performance improves further, indicating a synergistic effect.

\vspace{0.5em}\noindent\textbf{Image feature injection vs. AdaIN.} To analyze the impact of image feature injection and AdaIN, we compare Variant \#4 (without image feature injection or AdaIN) with Variant \#5 (with image feature injection only) and \#6 (with AdaIN only). Using ViT-B/32, we observe that Variant \#5 outperforms Variant \#4 across all four benchmarks, with an average gap of 0.8\%. While Variant \#6 shows a smaller improvement of 0.4\%, it outperforms Variant \#5 on ARO and SugarCrepe. These results suggest that image feature injection is more effective for improving compositionality than AdaIN, although the two methods are somewhat complementary. Through the improvements obtained using ViT-L/14 are a little bit different, both techniques improve performance over Variant \#4. Combining both methods in Variant \#7 leads to further improvements, with a more than 1\% increase in average accuracy for ViT-B/32 and a 0.65\% increase for ViT-L/14 over Variant \#4. This highlights the importance of reducing unintended changes in synthetic images, as such changes could cause the model to rely on them for distinguishing positive and negative samples rather than the semantic differences specified in the corresponding captions. Examples of synthetic images in Figure \ref{fig:syn-image-inject-1} and \ref{fig:syn-image-inject-2} in the Appendix show how image feature injection helps mitigate these unintended changes.

\vspace{0.5em}\noindent\textbf{Effects of the margin loss and adaptive margin strategies.} 
We compare three variants with distinct margin approaches: \emph{(1) Fixed (Variant \#8)}: a fixed margin without adaptive adjustments; \emph{(2) Adaptive (Variant \#9)}: the margin is calculated according to Eq. \eqref{eq:ada-margin}; and \emph{(3) Adaptive Inversed (Variant \#10)}: a larger margin is applied to samples with higher similarity differences, representing an inverse version of Eq. \eqref{eq:ada-margin} that prioritizes learning from easier samples. Comparing Variant \#8 with Variant \#7, we observe that \#8 improves by over 1\% for ViT-B/32 and 0.6\% for ViT-L/14 in average accuracy, suggesting that margin loss effectively aids in compositional understanding. Between Variants \#8, \#9, and \#10, we find that Variant \#9 outperforms Variant \#8 across all four benchmarks, with an average boost of over 0.3\% for both ViT-B/32 and ViT-L/14. In contrast, Variant \#10 results in a performance decline on three benchmarks and no improvement in average accuracy. These findings highlight the superior effectiveness of adaptive margins.

\vspace{0.5em}\noindent\textbf{Is applying adaptive margin loss only to hard samples sufficient?} To explore whether applying margin loss solely to hard samples is adequate, we construct Variant \#11, which uses only the positive set $\mathbb{P}$ and the hard negative set $\mathbb{N}_h$ for margin loss calculation in Eq. \eqref{eq:ada-margin-loss}, which is similar to learning strategies in \cite{zhang2023contrasting,lai2024improving}. When comparing Variant \#11 with Variant \#8 and \#9, we observe that Variant \#11 performs better on ARO and VL-CheckList, but worse on SugarCrepe and SugarCrepe++. We hypothesize that by focusing exclusively on the positive and hard negative sets, the model may learn nonsensical artifacts present in the hard negatives. These artifacts improve performance on ARO and VL-CheckList, where the test samples exhibit similar patterns, but hinder performance on SugarCrepe and SugarCrepe++, which are designed to avoid such patterns \cite{hsieh2023sugarcrepe}. In contrast, Variant \#9, which incorporates all sets in Eq. \eqref{eq:ada-margin-loss} for margin loss calculation, achieves better results on SugarCrepe and SugarCrepe++ with only a minor drop in performance on ARO and VL-CheckList. This suggests that easy negative samples serve as regularization, preventing overfitting to the artifacts in hard negative samples.

\begin{table}[tb]
\setlength{\tabcolsep}{1pt}
\begin{center}
\caption{Performance (\%) of \sysname{} using different subsets of training samples. ``Neg.'' represents synthetic negative samples, ``Pos.'' represents synthetic positive samples, and ``Prop.'' represents the proportion of training samples.}
\label{tab:abl-res-sample}
\scriptsize
\begin{tabular}{@{}cp{0.001cm}cccp{0.001cm}cccccp{0.001cm}@{}}
\toprule
Variant && Neg. & Pos. & Prop. && ~~ARO~~ & VL-CheckList & SugarCrepe & SugarCrepe++ & Average \\
\midrule
\#7 && \ding{52} & \ding{56} & 100\% && 72.12 & 74.65 & 86.41 & 57.99 & 72.79 \\
\#7 && \ding{52} & \ding{52} & 100\% && 74.12 & 76.35 & 85.40 & 66.44 & 75.58 \\
\midrule
\#9 && \ding{52} & \ding{52} & 20\% && 75.90 & 77.56 & 85.78 & 64.44 & 75.92 \\
\#9 && \ding{52} & \ding{52} & 50\% && 77.90 & 79.38 & 86.71 & 64.21 & 77.05 \\
\#9 && \ding{52} & \ding{52} & 100\% && 77.15 & 79.16 & 87.11 & 66.12 & 77.38 \\
\bottomrule
\end{tabular}
\end{center}
\vspace{-1.0em}
\end{table}

\vspace{0.5em}\noindent\textbf{Effects of synthetic positive samples.} We compare the performance of training with only synthetic negative samples versus using both synthetic negative and positive samples in Table \ref{tab:abl-res-sample}. Based on Variant \#7, we train ViT-B/32 with these two combinations of training data. We observe nearly 3\% decrease in average accuracy when using only synthetic negative samples, compared to using both synthetic negative and positive samples. This decrease is particularly noticeable on SugarCrepe++, ARO, and VL-CheckList. These results underscore the importance of incorporating synthetic positive samples in training, as they provide essential information about variations that maintain semantic consistency, which is critical for compositional understanding.

\vspace{0.5em}\noindent\textbf{Influence of model size and training data size.} To evaluate the impact of model size on compositional understanding, we compare the performance of ViT-B/32 and ViT-L/14 in Table \ref{tab:abl-res-all}. We find that ViT-L/14 only provides a modest improvement of about 0.5\% of average accuracy over ViT-B/32. This suggests that increasing model size does not significantly enhance compositional understanding. To investigate the effect of training data size, we train ViT-B/32 using 20\% and 50\% of randomly sampled training data and report the results in Table \ref{tab:abl-res-sample}. We observe that performance improves as the proportion of training data increases, indicating that a larger training set helps the model better capture compositional knowledge. However, the performance gain diminishes when the training data size increases from 50\% to 100\%. Perhaps generating or selecting high-quality data would be more effective than simplistic data scaling.

Additional experiments on hyperparameter analysis can be found in Appendix \ref{supp-sec:res}.

\section{Conclusion}
In this paper, we introduce \sysname{} to enhance the compositional understanding capabilities of VLMs by generating and training with synthetic data. To tackle two key challenges in using synthetic data---namely, the difficulty of generating accurate variations and the inconsistency in cross-modal alignment quality---\sysname{} integrates image feature injection into a T2I model to improve the quality of synthetic variations and introduces an adaptive margin loss to account for varying levels of cross-modal alignment for effectively learning nuanced distinctions. Experiments on four visual-language compositional understanding benchmarks demonstrate the effectiveness of \sysname{}.

\section{Acknowledgments}
This work has been supported by the Nanyang Associate Professorship and the National Research Foundation Fellowship (NRF-
NRFF13-2021-0006), Singapore.
Any opinions, findings, conclusions,
or recommendations expressed in this material are those of the authors and do not reflect the views of the funding agencies.


\clearpage
\renewcommand{\thesection}{A\arabic{section}}
\renewcommand{\theHsection}{A\arabic{section}}
\setcounter{section}{0}
\renewcommand{\thefigure}{A\arabic{figure}}
\setcounter{figure}{0}
\renewcommand{\thetable}{A\arabic{table}}
\setcounter{table}{0}
\maketitlesupplementary

\noindent
The Appendix is organized as follows:
\begin{itemize}
    \item Section \ref{supp-sec:method} presents additional details about \sysname{}.
    \item Section \ref{supp-sec:exp} provides further details on the experimental setup.
    \item Section \ref{supp-sec:res} includes additional experimental results.
\end{itemize}

\section{Details of \sysname{}}\label{supp-sec:method}
The prompts used as input to the LLM for generating negative and positive captions are presented in Figure \ref{fig:prompts}.

\begin{figure}[!htb]
    \centering
    \captionsetup[subfigure]{skip=0pt}
    \begin{subfigure}{\linewidth}
        \centering
        \begin{tcolorbox}[colback=gray!10, colframe=black, boxrule=0.5pt, arc=2mm, fontupper=\small]
            You are an assistant assigned to help a human user edit a given sentence that describes an image. Make a minor change to the sentence by randomly altering, omitting, inserting, or replacing one word or phrase. Although the change should be minor, it must result in a significant difference in the sentence's meaning, making it unable to describe the original image. Use the provided template and respond with a single, valid sentence.
            
            User: \{\}
            
            Assistant: Sure! Here's my edit: 
        \end{tcolorbox}
        \caption{Prompts used to generate negative captions.}
        \label{fig:prompt-neg}
    \end{subfigure}
    
    \vspace{5pt}

    \begin{subfigure}{\linewidth}
        \centering
        \begin{tcolorbox}[colback=gray!10, colframe=black, boxrule=0.5pt, arc=2mm, fontupper=\small]
            You are an assistant assigned to help a user edit a sentence that describes an image. Make a minor change to the sentence by randomly altering, omitting, inserting, or replacing one word or phrase. The new sentence must strictly retain the same meaning as the original sentence. Use the provided template and respond with a single, valid sentence.
            
            User: \{\}
            
            Assistant: Sure! Here's my edit: 
        \end{tcolorbox}
        \caption{Prompts used to generate positive captions.}
        \label{fig:prompt-pos}
    \end{subfigure}

    \caption{Prompts used to generate negative and positive captions.}
    \label{fig:prompts}
\end{figure}

\section{Experimental Setup}\label{supp-sec:exp}
\noindent\textbf{Data Synthesis.} For caption generation, we utilize the Llama-2-Chat 13B model\footnote{\scriptsize{\url{https://huggingface.co/meta-llama/Llama-2-13b-chat}}}, with the temperature set to 0.9, top-k set to 100, and top-p set to 0.9 for sampling. For image generation, we use the LCM model\footnote{\scriptsize{\url{https://huggingface.co/SimianLuo/LCM_Dreamshaper_v7}}} for its swift inference with few steps \cite{luo2023latent}. The pretrained CLIP ViT-L/14 \cite{radford2021learning} is used as the image feature extractor for injecting image features. We perform 8 inference steps with LCM to generate each image.

\vspace{0.3em}\noindent\textbf{Hyperparameter Selection.} First, we use only real training samples to select $\tau$ and $b$. The optimal values are determined by searching for the ones that minimize the training loss at the first training step, aiming to preserve the output distribution from the pretrained model. After searching, we set $\tau = 0.01$ and $b = -30.0$. Next, we select the base learning rate, weight decay, and LoRA adapter rank based on performance on the COCO-2014 validation set, in which the model is trained exclusively on real samples. According to the performance on the validation set, these hyperparameter are set to a base learning rate of 0.01, weight decay of 0.5, and LoRA adapter rank of 16. Then, we construct a validation set composed of the CIFAR-10 \cite{krizhevsky2009learning} test set and a randomly selected 5\% of samples from ARO-Attribute and ARO-Relation, to balance the performance on coarse-grained and fine-grained tasks. Using this validation set, we train the model on both real and synthetic samples and use the validation performance to determine the remaining hyperparameters: $m_0$, $\alpha$, $\beta$, $\gamma$ and $\lambda$. The effects of these hyperparameters are shown in Table \ref{tab:res-hyperparameters}.

\section{Experimental Results}\label{supp-sec:res}
\noindent\textbf{Performance on each subset of the four benchmarks.} Table \ref{tab:res-aro-vl}, \ref{tab:res-sugarcrepe} and \ref{tab:res-sugarcrepepp} present the performance of different methods on each subset of the four benchmarks.

\begin{table*}[t]
\setlength{\tabcolsep}{4pt}
\begin{center}
\caption{Comparison of accuracy (\%) between \sysname{} and baselines on ARO and VL-CheckList. ``img'' represents images, ``cap'' represents captions, ``syn'' represents synthetic data.}
\label{tab:res-aro-vl}
\footnotesize
\begin{tabular}{@{}lp{0.01cm}cccccp{0.01cm}cccp{0.01cm}cccc@{}}
\toprule
\multirow{4}{*}{Method} && \multicolumn{5}{c}{Training Data} && \multicolumn{3}{c}{\multirow{3}{*}{ARO}} && \multicolumn{4}{c}{\multirow{3}{*}{VL-CheckList}} \\
\cmidrule{3-7}
~ && \multirow{2}{*}{Source} & \multirow{2}{*}{\makecell{\# real\\img}} & \multirow{2}{*}{\makecell{\# real\\cap}} & \multirow{2}{*}{\makecell{\# syn\\img}} & \multirow{2}{*}{\makecell{\# syn\\cap}} && ~ & ~ & ~ && ~ & ~ & ~ & ~ \\
\cmidrule{9-11}\cmidrule{13-16}
~ && ~ & ~ & ~ & ~ & ~ && Relation & Attribute & Average && Attribute & Relation & Object & Average \\
\midrule
CLIP-ZeroShot\cite{radford2021learning} && - & - & - & - & - && 59.22 & 62.86 & 61.03 && 67.05 & 66.71 & 85.72 & 73.16 \\
\midrule
CLIP-Finetune\cite{radford2021learning} && COCO & 82K & 410K & 0 & 0 && 63.02 & 65.16 & 64.09 && 66.74 & 64.43 & 86.86 & 72.78  \\
SDS-CLIP \cite{basu2023augmenting} && COCO & 82K & 410K & 0 & 0 && 53.0 & 62.0 & 57.5 && - & - & - & - \\
\cite{sahin2024enhancing} && COCO & 0 & 0 & 82K & 82K && - & - & - && 70.7 & 53.8 & 85.1 & 69.87 \\
AMR-NegCLIP \cite{shou2024enhancing} && COCO & 100K & 100K & 0 & 500K && 83.2 & 75.6 & 79.4 && - & - & - & - \\
NegCLIP \cite{yuksekgonul2022and} && COCO & 100K & 100K & 0 & 500K && 81.0 & 71.0 & 76.0 && 70.9 & 68.9 & 84.1 & 74.6 \\
MosaiCLIP \cite{singh2023coarse} && COCO & 109K & 109K & 0 & 981K && 82.6 & 78.0 & 80.3 && 70.1 & 71.3 & 89.0 & 76.8 \\
FSC-CLIP \cite{oh2024preserving} && COCO & 100K & 100K & 0 & 1.5M && - & - & - && - & - & - & 77.20 \\
CE-CLIP \cite{zhang2023contrasting} && COCO & 82K & 410K & 0 & 2M && 83.00 & 76.40 & 79.70 && 72.62 & 71.75 & 84.65 & 76.34 \\
COMO \cite{lai2024improving} && COCO & 113K & 567K & 567K & 567K && - & - & - && 73.44 & 71.16 & 86.20 & 76.93 \\
\midrule
\sysname{} && COCO & 82K & 410K & 820K & 820K && 80.10 & 74.19 & 77.15 && 73.72 & 72.99 & 90.76 & 79.16 \\
\midrule
SPEC \cite{peng2023synthesize} && LAION & 20K & 20K & 20K & 20K && 73.7 & 66.4 & 70.1 && - & - & - & - \\
\cite{doveh2023teaching} && CC3M & 3M & 3M & 0 & 9M && - & - & - && 71.97 & 68.95 & 85.00 & 75.31 \\
CE-CLIP+ \cite{zhang2023contrasting} && COCO+CC3M & 3M & 3M & 0 & 15M && 83.6 & 77.1 & 80.35 && 76.76 & 74.70 & 86.30 & 79.25 \\
CLOVE \cite{castro2024clove} && LAION-COCO & $>$1B & $>$1B & 0 & $>$1B && 69.0 & 77.4 & 73.2 && - & - & - & - \\
syn-CLIP \cite{cascante2023going} && SyViC & 0 & 0 & $>$1M & $>$1M && 71.40 & 66.94 & 69.17 && 70.37 & 69.39 & 84.75 & 74.84 \\
FiGCLIP \cite{khan2024figclip} && VidSitu & \multicolumn{2}{c}{20K videos} & 0 & 0 && 68.01 & 65.99 & 67.00 && - & - & - & - \\
\bottomrule
\end{tabular}
\end{center}
\end{table*}

\begin{table*}[t]
\setlength{\tabcolsep}{3pt}
\begin{center}
\caption{Comparison of accuracy (\%) between \sysname{} and baselines on SugarCrepe. ``img'' represents images, ``cap'' represents captions, ``syn'' represents synthetic data.}
\label{tab:res-sugarcrepe}
\footnotesize
\begin{tabular}{@{}lp{0.01cm}cccccp{0.01cm}ccp{0.01cm}cccp{0.01cm}ccp{0.01cm}c@{}}
\toprule
\multirow{4}{*}{Method} && \multicolumn{5}{c}{Training Data} && \multicolumn{2}{c}{\multirow{3}{*}{Add}} && \multicolumn{3}{c}{\multirow{3}{*}{Replace}} && \multicolumn{2}{c}{\multirow{3}{*}{Swap}} && \multirow{4}{*}{Average} \\
\cmidrule{3-7}
~ && \multirow{2}{*}{Source} & \multirow{2}{*}{\makecell{\# real\\img}} & \multirow{2}{*}{\makecell{\# real\\cap}} & \multirow{2}{*}{\makecell{\# syn\\img}} & \multirow{2}{*}{\makecell{\# syn\\cap}} && ~ & ~ && ~ & ~ & ~ && ~ & ~ \\
\cmidrule{9-10}\cmidrule{12-14}\cmidrule{16-17}
~ && ~ & ~ & ~ & ~ & ~ && Attribute & Object && Attribute & Object & Relation && Attribute & Object && ~ \\
\midrule
CLIP-ZeroShot\cite{radford2021learning} && - & - & - & - & - && 69.22 & 77.40 && 80.33 & 90.98 & 69.49 && 64.71 & 61.63 && 73.39 \\
\midrule
CLIP \cite{radford2021learning} (Finetune) && COCO & 82K & 410K & 0 & 0 && 78.03 & 88.12 && 85.79 & 93.58 & 73.83 && 71.77 & 68.29 && 79.92 \\
AMR-NegCLIP \cite{shou2024enhancing} && COCO & 100K & 100K & 0 & 500K && - & - && - & - & - && - & - && 79.92 \\
NegCLIP \cite{yuksekgonul2022and} && COCO & 100K & 100K & 0 & 500K && 82.80 & 88.80 && 85.91 & 92.68 & 76.46 && 75.38 & 75.20 && 82.46 \\
FSC-CLIP \cite{oh2024preserving} && COCO & 100K & 100K & 0 & 1.5M && - & - && - & - & - && - & - && 85.10 \\
CE-CLIP \cite{zhang2023contrasting} && COCO & 82K & 410K & 0 & 2M && 93.4 & 92.4 && 88.8 & 93.1 & 79.0 && 77.0 & 72.8 && 85.2 \\
\midrule
\sysname{} && COCO & 82K & 410K & 820K & 820K && 93.49 & 92.43 && 88.95 & 95.82 & 78.94 && 81.38 & 78.77 && 87.11  \\
\midrule
CounterCurate \cite{zhang2024countercurate} && Flickr & 30K & 30K & 150K & 150K && 86.71 & 90.35 && 87.94 & 95.94 & 76.24 && 73.57 & 68.57 && 82.76 \\
CE-CLIP+ \cite{zhang2023contrasting} && COCO+CC3M & 3M & 3M & 0 & 15M && 94.9 & 93.8 && 90.8 & 93.8 & 83.2 && 79.3 & 76.8 && 87.5 \\
CLOVE \cite{castro2024clove} && LAION-COCO & $>$1B & $>$1B & 0 & $>$1B && - & - && - & - & - && - & - && 79.92 \\
IL-CLIP \cite{zheng2024iterated} && CC12M & 12M & 12M & 0 & 0 && - & - && - & - & - && - & - && 70.34 \\
SF-CLIP \cite{sameni2024building} && YFCC15M & 15M & 15M & 0 & 0 && - & - && - & - & - && - & - && 71.20 \\
FiGCLIP \cite{khan2024figclip} && VidSitu & \multicolumn{2}{c}{20K videos} & 0 & 0 && 72.5 & 77.4 && 81.1 & 91.8 & 69.4 && 66.1 & 63.8 && 74.6 \\
\bottomrule
\end{tabular}
\end{center}
\end{table*}

\begin{table*}[t]
\begin{center}
\caption{Comparison of accuracy (\%) between \sysname{} and baselines on SugarCrepe++. ``img'' represents images, ``cap'' represents captions, ``syn'' represents synthetic data.}
\label{tab:res-sugarcrepepp}
\footnotesize
\begin{tabular}{@{}lp{0.01cm}cccccp{0.01cm}cccp{0.01cm}ccp{0.01cm}c@{}}
\toprule
\multirow{4}{*}{Method} && \multicolumn{5}{c}{Training Data} && \multicolumn{3}{c}{\multirow{3}{*}{Replace}} && \multicolumn{2}{c}{\multirow{3}{*}{Swap}} && \multirow{4}{*}{Average} \\
\cmidrule{3-7}
~ && \multirow{2}{*}{Source} & \multirow{2}{*}{\makecell{\# real\\img}} & \multirow{2}{*}{\makecell{\# real\\cap}} & \multirow{2}{*}{\makecell{\# syn\\img}} & \multirow{2}{*}{\makecell{\# syn\\cap}} && ~ & ~ & ~ && ~ & ~ \\
\cmidrule{9-11}\cmidrule{13-14}
~ && ~ & ~ & ~ & ~ & ~ && Attribute & Object & Relation && Attribute & Object && ~ \\
\midrule
CLIP-ZeroShot\cite{radford2021learning} && - & - & - & - & - && 65.61 & 86.80 & 56.26 && 45.21 & 45.18 && 59.81 \\
\midrule
CLIP-Finetune\cite{radford2021learning} && COCO & 82K & 410K & 0 & 0 && 69.03 & 90.61 & 56.33 && 49.24 & 46.21 && 62.27 \\
NegCLIP\cite{yuksekgonul2022and} && COCO & 100K & 100K & 0 & 500K && 69.41 & 89.53 & 52.27 && 57.99 & 55.25 && 64.89 \\
\midrule
\sysname{} && COCO & 82K & 410K & 820K & 820K && 68.90 & 89.76 & 52.34 && 57.95 & 61.63 && 66.12 \\
\midrule
\cite{doveh2023teaching} && CC3M & 3M & 3M & 0 & 9M && 56.98 & 80.93 & 47.30 && 48.4 & 42.98 && 55.32 \\
\bottomrule
\end{tabular}
\end{center}
\end{table*}

\begin{figure}[!ht]
\centering
\includegraphics[width=\linewidth]{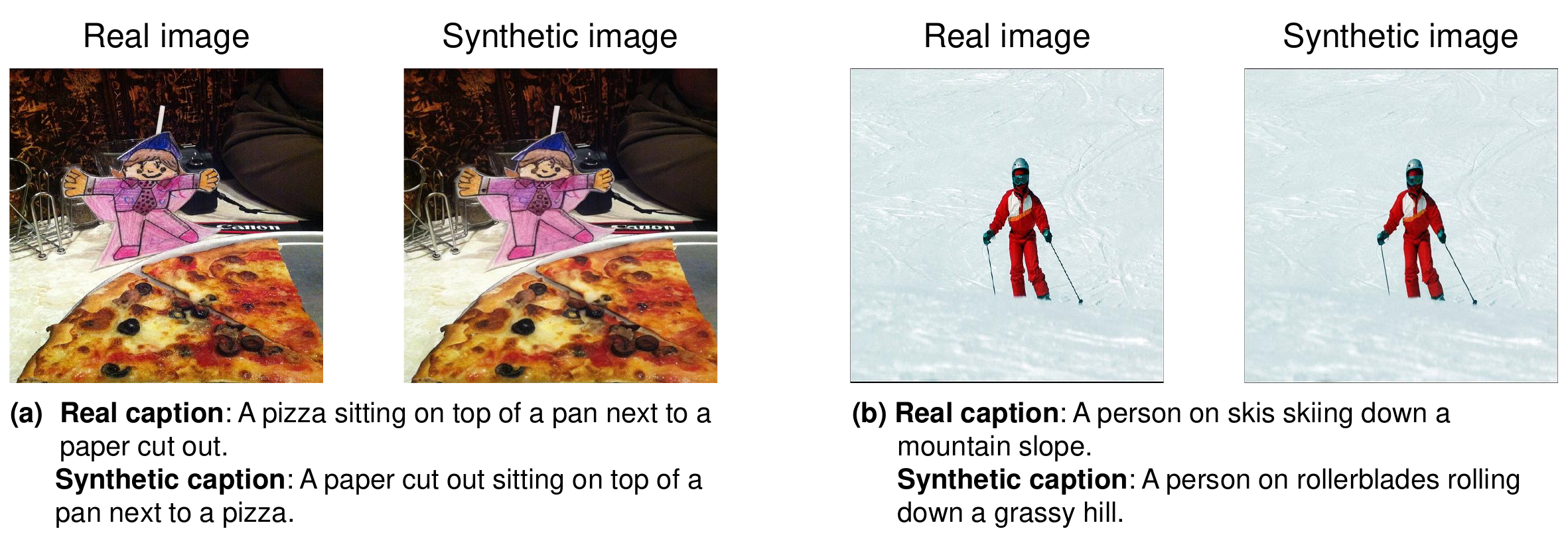}
\caption{Examples of synthetic samples from StyleAligned. The algorithm did not alter the image content according to the caption. }
\label{fig:syn-stylealigned}
\end{figure}

\vspace{0.5em}\noindent\textbf{Comparison with other images generation methods.} We compare our image generation method with StyleAligned \cite{hertz2024style}. For a fair comparison, we use an ablated version \#7 of \sysname{} (Sec. \ref{subsec:ablation}, main paper) without the adaptive margin loss. Both methods use synthetic captions that we generate. As shown in Table \ref{tab:comp-stylealigh}, StyleAligned performs about 1\% worse than our method on the four compositional benchmarks, which illustrates the effectiveness of image feature injection in \sysname{}. In Figure \ref{fig:syn-stylealigned}, we show two synthetic images from StyleAligned, where it fails to align the generated content with the synthetic captions. We hypothesize that the diffusion trajectory of the real image imposes strong constraints on the image generation model, making StyleAligned difficult to edit the image to match the synthetic caption. This issue is similar to the zero-shot image editing methods \cite{meng2021sdedit,brooks2023instructpix2pix,garibi2024renoise}, which provide incorrect guidance during model training and lead to limited improvements on compositional understanding tasks. Moreover, StyleAligned requires DDIM inversion to obtain the inverted diffusion trajectory from the real image, making it computationally expensive and impractical for large-scale image generation.

\begin{table}[!ht]
\setlength{\tabcolsep}{2pt}
\begin{center}
\caption{Performance comparison (\%) between \sysname{} and StyleAligned.}
\label{tab:comp-stylealigh}
\scriptsize
\begin{tabular}{@{}cp{0.001cm}cccccp{0.001cm}@{}}
\toprule
Variant && ~~ARO~~ & VL-CheckList & SugarCrepe & SugarCrepe++ & Average \\
\midrule
StyleAligned \cite{hertz2024style} && 72.60 & 75.03 & 85.70 & 65.25 & 74.65 \\
\sysname{} (\#7) && 74.12 & 76.35 & 85.40 & 66.44 & 75.58 \\
\bottomrule
\end{tabular}
\end{center}
\end{table}

\begin{figure*}[t]
\centering
\includegraphics[width=\linewidth]{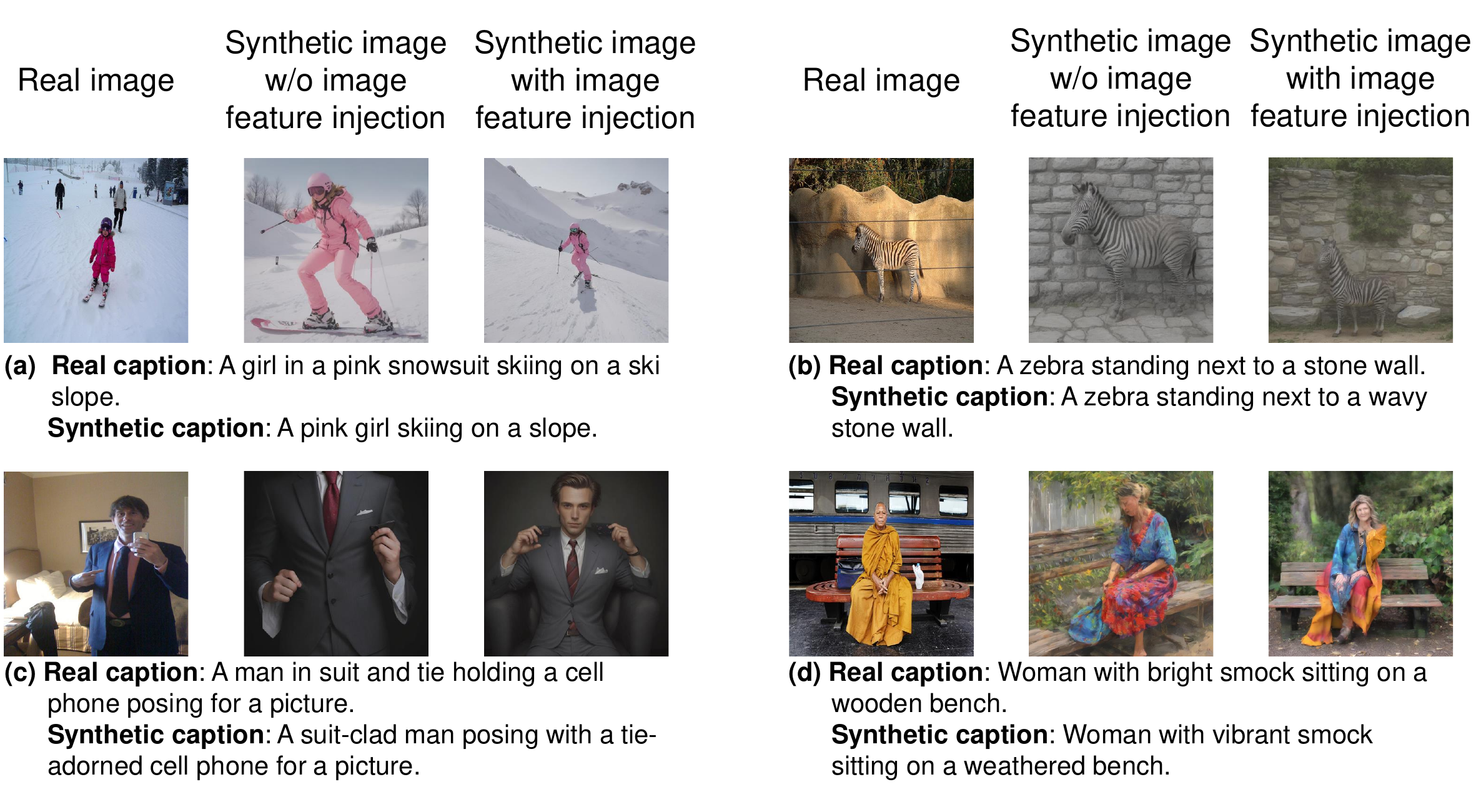}
\caption{Examples of synthetic samples without and with image feature injection. In these examples, the image feature injection technique achieves alignment of the subject size and the viewing angle with those in real images.
}
\label{fig:syn-image-inject-1}
\end{figure*}

\begin{figure*}[t]
\centering
\includegraphics[width=\linewidth]{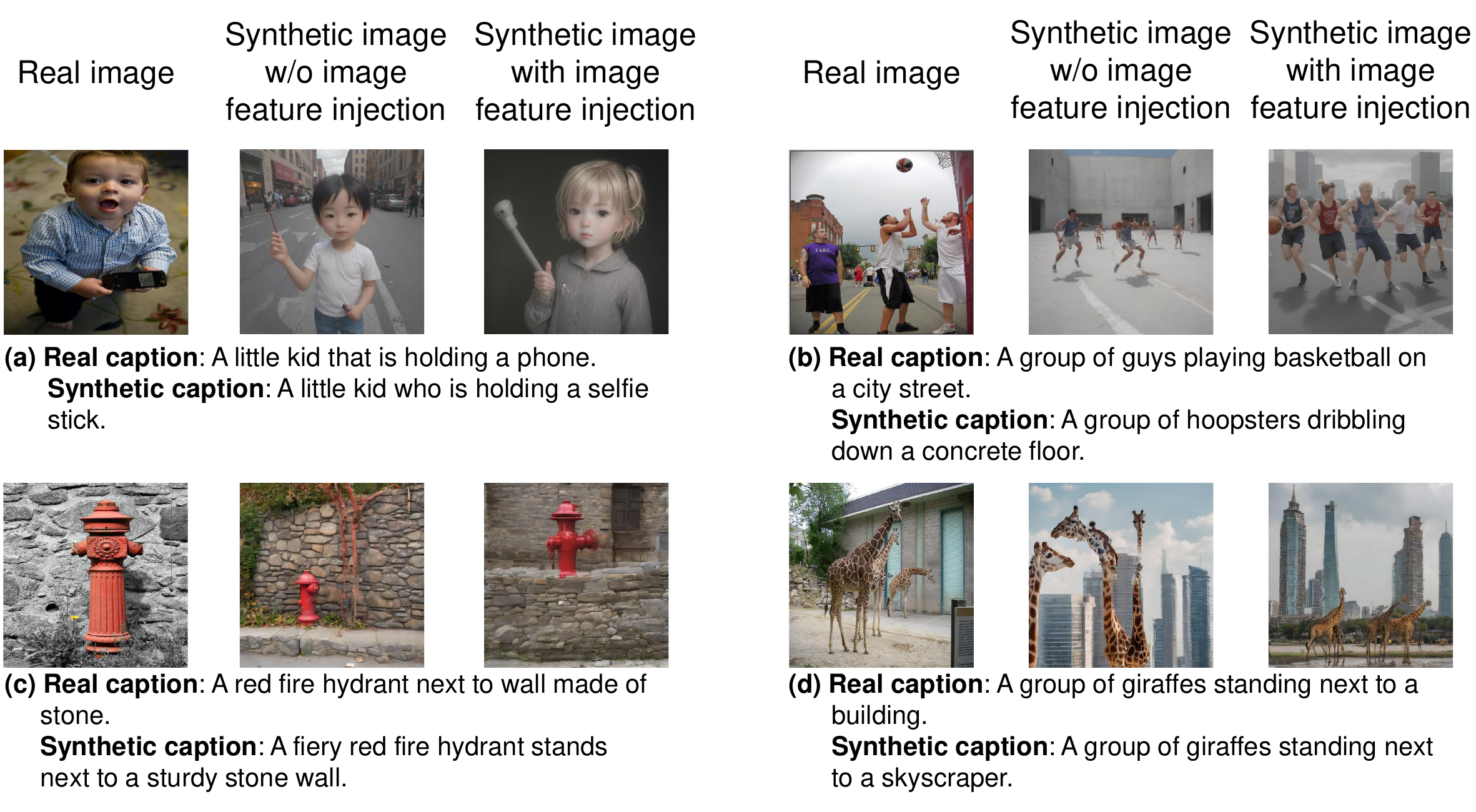}
\caption{Examples of synthetic samples without and with image feature injection. In these examples, the image feature injection primarily helps to generate backgrounds that resemble those in real images. For example, in (d), both the first and the third images show the ground, whereas the second image does not.}
\label{fig:syn-image-inject-2}
\end{figure*}

\vspace{0.5em}\noindent\textbf{Effects of image feature injection.} In Figure \ref{fig:syn-image-inject-1} and \ref{fig:syn-image-inject-2}, we present examples of synthetic images to illustrate how image feature injection helps mitigate unintended changes. In Figure \ref{fig:syn-image-inject-1}, we observe that feature injection helps to generate images with similar object size and viewing angle to the real image. For example, in (a), the real image depicts a wide shot of a girl, while the synthetic image without feature injection produces a close-up shot despite aligning with the caption. With feature injection, the synthetic image maintains a wide shot, resembling the real image. Similar effects are seen in (b) and (c). In (d), the synthetic image with feature injection preserves the viewing angle of the real image, whereas the one without feature injection deviates from it. In Figure \ref{fig:syn-image-inject-2}, we observe that feature injection helps generate backgrounds that resemble the real image. For example, in (a), the real image and the synthetic image without feature injection depicts an outdoor street scene, creating a noticeable difference. With feature injection, the single-colored background makes the synthetic image more similar to the real one. In (b), the sky occupies much of background in the real image as well as the image generated with feature injection, whereas the one without feature injection shows little sky. Also, the basketball is present in both the real and the synthetic image with feature injection but not in the middle image. Similar effects are observed in (c) and (d). These examples show that image feature injection reduces unintended variations not captured by the caption, enhancing the usefulness of synthetic samples for training VLMs.

\begin{table}[t]
\setlength{\tabcolsep}{2pt}
\begin{center}
\caption{Performance of \sysname{} with different hyperparameters. ``ARO-Rel'' refers to the ARO-Relation validation subset, and ``ARO-Att'' refers to the ARO-Attribute validation subset, both consisting of a randomly selected 5\% of the full set, as described in Sec. \ref{supp-sec:exp}.} 
\label{tab:res-hyperparameters}
\scriptsize
\begin{tabular}{@{}cccccp{0.01cm}ccccp{0.01cm}c@{}}
\toprule
\multirow{2}{*}{$\lambda$} & \multirow{2}{*}{$\alpha$} & \multirow{2}{*}{$m_0$} & \multirow{2}{*}{$\beta$} & \multirow{2}{*}{$\gamma$} && \multicolumn{4}{c}{Validation} && \multirow{2}{*}{\makecell{Test\\Average}} \\
\cmidrule{7-10}
~ & ~ & ~ & ~ & ~ && CIFAR-10 & ARO-Rel & ARO-Att & Average && ~ \\
\midrule
0.0 & 0.0 & - & - & - && 86.56 & 78.79 & 76.52 & 80.62 && 75.58 \\
\midrule
0.001 & 0.0 & 0.01 & 0.0 & 0.0 && 85.02 & 78.21 & 72.72 & 78.65 && 75.95 \\
0.01 & 0.0 & 0.01 & 0.0 & 0.0 && 83.66 & 81.77 & 76.46 & 80.63 && 76.78 \\
0.1 & 0.0 & 0.01 & 0.0 & 0.0 && 85.02 & 81.29 & 78.94 & 80.47 && 76.94 \\
\midrule
0.01 & 1.0 & 0.01 & 0.0 & 0.0 && 83.18 & 81.10 & 76.52 & 80.27 && 77.21 \\
0.01 & 10.0 & 0.01 & 0.0 & 0.0 && 86.46 & 81.89 & 75.05 & 81.13 && 77.27 \\
0.01 & 100.0 & 0.01 & 0.0 & 0.0 && 87.64 & 74.93 & 75.19 & 79.25 && 75.28 \\
\midrule
0.01 & 10.0 & 0.005 & 0.0 & 0.0 && 85.91 & 79.87 & 78.76 & 81.51 && 77.08 \\
0.01 & 10.0 & 0.01 & 0.0 & 0.0 && 86.46 & 81.89 & 75.05 & 81.13 && 77.27 \\
0.01 & 10.0 & 0.02 & 0.0 & 0.0 && 84.85 & 79.41 & 75.28 & 79.85 && 76.79 \\
\midrule
0.01 & 10.0 & 0.005 & -0.02 & 1.0 && 86.46 & 80.08 & 78.90 & 81.81 && 77.38 \\
0.01 & 10.0 & 0.005 & -0.03 & 1.0 && 86.75 & 81.08 & 76.43 & 81.42 && 77.25 \\
0.01 & 10.0 & 0.005 & -0.02 & 3.0 && 87.31 & 80.79 & 76.52 & 81.54 && 77.23 \\
\bottomrule
\end{tabular}
\end{center}
\end{table}

\vspace{0.5em}\noindent\textbf{Effects of hyperparameters.} Table \ref{tab:res-hyperparameters} presents the performance of \sysname{} with different hyperparameter settings. For $\lambda$, we observe that $\lambda=0.01$ achieves the highest average validation accuracy, leading us to select it for subsequent experiments. Similarly, for $\alpha$, the best performance is obtained with $\alpha=10.0$, which is used in other experiments. When evaluating different values of $m_0$, we find that $m_0=0.005$ yields the best results. Finally, we examine various combinations of $\beta$ and $\gamma$ and observe that $\beta=-0.02$ and $\gamma=1.0$ provide the best validation performance. Thus, this combination is selected as the optimal hyperparameter setting.

%

\FloatBarrier
\clearpage
{
    \small
    \bibliographystyle{ieeenat_fullname}
    \bibliography{main}
}

\end{document}